\documentclass[runningheads]{llncs}
\usepackage{graphicx}
\usepackage{url}
\begin{document}
\title{Unsupervised bias discovery\\ in medical image segmentation}
\titlerunning{Unsupervised bias discovery in medical image segmentation}
%
\author{Nicol\'as Gaggion \and
Rodrigo Echeveste \and
Lucas Mansilla \and
Diego H. Milone \and
Enzo Ferrante}

\authorrunning{N. Gaggion et al.}

\institute{Research Institute for Signals, Systems and Computational Intelligence, sinc(i)\\ (CONICET, Universidad Nacional del Litoral)\\ Santa Fe, Argentina}

\maketitle              
\begin{abstract}
It has recently been shown that deep learning models for anatomical segmentation in medical images can exhibit biases against certain sub-populations defined in terms of protected attributes like sex or ethnicity. In this context, auditing fairness of deep segmentation models becomes crucial. However, such audit process generally requires access to ground-truth segmentation masks for the target population, which may not always be available, especially when going from development to deployment. Here we propose a new method to anticipate model biases in biomedical image segmentation in the absence of ground-truth annotations. Our unsupervised bias discovery method leverages the reverse classification accuracy framework to estimate segmentation quality. Through numerical experiments in synthetic and realistic scenarios we show how our method is able to successfully anticipate fairness issues in the absence of ground-truth labels, constituting a novel and valuable tool in this field. 

\end{abstract}
\section{Introduction}
An ever growing body of work has shown that machine learning systems can be systematically biased against certain sub-populations based on attributes like race or gender in a variety of settings \cite{ricci2022addressing}. When it comes to machine learning systems analyzing health data, this topic becomes extremely relevant, particularly in medical image computing (MIC) tasks like computed assisted diagnosis \cite{larrazabal2020gender} or anatomical segmentation \cite{puyol2021fairness}. It is hence vital to audit the models considering fairness metrics to assess potential disparate performance of various types among subgroups \cite{liu2022medical}. A usual way to detect such fairness issues consists in employing the standard \textit{group fairness} framework to audit bias in machine learning models. This framework usually requires access to a dataset of images with two important pieces of information: the \textit{demographic attributes} used to define the groups of analysis, and the \textit{ground truth labels} for the task of interest. However, in many situations, we may not have access to them. Medical images, for example, may lack the necessary metadata due to privacy concerns. Moreover, obtaining expert label annotations for extensive image databases can be a time-consuming and costly endeavor, significantly limiting the availability of these annotations. More importantly, in real life scenarios, we may be interested in auditing existing deployed systems in a new target population for which we do not have ground-truth labels to ensure these models are still fair. To tackle these issues, here we propose an unsupervised bias discovery (UBD) method, which to our knowledge is the first one specifically designed for biomedical image segmentation. 

The term UBD refers to the process of identifying and uncovering biases in a machine learning system without using ground-truth labels or demographic information. Two primary UBD scenarios can be considered. The first one, when we have access to ground-truth (such as classification labels or anatomical masks for image segmentation) but lack knowledge about the specific sub-population metadata for sub-group analysis. Methods addressing this issue have been recently proposed in fairness literature for machine learning \cite{krishnakumar2021udis,lahoti2020fairness}. The second scenario, of main interest to us, appears when we possess demographic attributes at an individual level to construct pre-defined analysis groups (e.g. gender, sex, age, ethnicity) but lack ground-truth annotations for the target population. Here we focus in this last case, i.e. performing UBD in the absence of ground-truth to anticipate fairness issues in unseen subjects, particularly in the context of medical image segmentation. Given recent research indicating how fairness properties may not transfer under distribution shifts in healthcare \cite{schrouff2022diagnosing}, understanding how to anticipate fairness issues in new populations with unforeseen distribution shifts, especially when ground-truth labels are unavailable, becomes of paramount importance.

\section{Related work}
Anatomical segmentation is a fundamental task in medical image analysis, with applications ranging from computational anatomy studies and patient follow-up, to radiotherapy planning \cite{fu2021review,cubero2023}. Several methods have been proposed in the literature to estimate segmentation performance in the absence of ground-truth annotations, usually in the context of automatic quality control pipelines. Some of these methods leverage the concept of predictive uncertainty \cite{czolbe2021segmentation,cubero2023} under the hypothesis that highly uncertain predictions will correlate with erroneous pixels. However, these methods heavily rely on the quality of the uncertainty estimates provided by the segmentation model. In \cite{fournel2021medical}, the authors proposed an alternative learning-based approach where a convolutional neural network (CNN) is trained to predict the Dice-Sørensen coefficient (or DSC, a commonly used metric for segmentation quality) from pairs of images with the corresponding predicted segmentation. This approach is agnostic to the segmentation model, but it does require to train an extra CNN to predict the DSC score. Here we build on top of a different approach introduced by Valindria et al \cite{valindria2017reverse}, entitled reverse classification accuracy (RCA). This method proposes to construct a \textit{reverse classifier}, using only one test image and its predicted segmentation as pseudo ground truth. Subsequently, this classifier is assessed on a reference dataset containing available segmentations, and its performance is used as a proxy of the quality of the predicted segmentation. Different variants of RCA were originally proposed, including atlas-forest, CNN-based RCA and atlas-based label propagation RCA via classic image registration. Here we implement our own deep learning variant of atlas-based label propagation RCA, which has shown to be fast and reliable in previous work \cite{mansilla2018segmentacion,mansilla2020learning} (see details in Section \ref{sec:ubd}), and use it to perform UBD. 

\paragraph{Contributions: } To the best of our knowledge, this is the first study to explore unsupervised bias discovery in biomedical image segmentation. To this end, we propose a deep learning variant of atlas-based label propagation RCA and demonstrate its capacity to reveal hidden biases in segmentation models that exhibit disparate performance across specific sub-populations. The proposed method is applied to the task of chest X-ray anatomical segmentation, revealing sex biases in situations where ground-truth masks are unavailable. This is fundamentally important when implementing anatomical segmentation systems in real and dynamic clinical scenarios, as it may help to monitor that fairness is maintained when models are deployed into production. 

\section{Unsupervised bias discovery via reverse classification accuracy}
\label{sec:ubd}
\begin{figure}[t!]
    \centering
    \includegraphics[width=\textwidth]{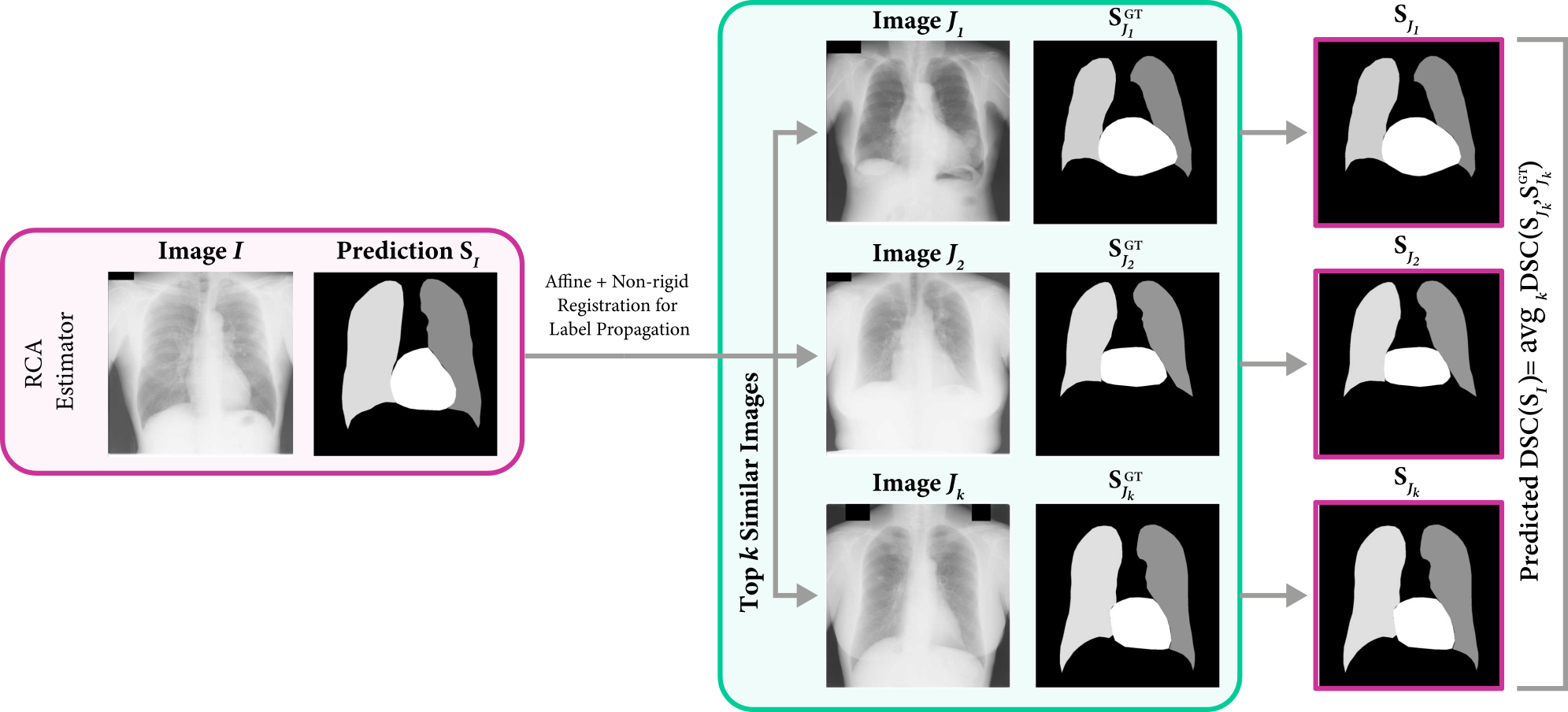}
    \caption{Outline of the RCA framework (based on \cite{valindria2017reverse}). The DSC-score for the predicted segmentation $S_I$ (for which we do not have ground-truth) is estimated by registering image $I$ to the top-$k$ similar images from the reference database, and computing the DSC-score of the propagated labels with respect to the corresponding segmentation masks. The mean DSC computed for the $k$ reference masks is then used as a DSC estimate of the predicted segmentation $S_I$.}
    \label{fig:rca}
\end{figure}

Let us briefly start by describing the atlas based label-propagation RCA framework \cite{valindria2017reverse} adopted in our work to estimate individual DSC coefficients per image. The process starts with an image $I$ and its corresponding predicted segmentation $S_{I}$ for which quality needs to be assessed (see Figure \ref{fig:rca}). We will refer to this image as the \textit{atlas}. Additionally, there is a reference database containing images $J_i$ with known ground-truth segmentation masks $S^{GT}_{J_i}$. 

Non-rigid registration is then applied to align the atlas with the reference images in the database, deforming it via a deformation field $D_i$ to match their anatomy. Through label propagation, the predicted segmentation mask under evaluation ($S_I$) is transferred to the corresponding target images by warping it with the resulting deformation fields, providing a candidate segmentation mask $\hat{S}_{J_i} = S_I \circ D_i$ per reference image $J_i$. Subsequently, we evaluate the quality of the propagated labels $\hat{S}_{J_i}$ by comparing them with the corresponding ground-truth segmentation masks $S^{GT}_{J_i}$ using DSC (or any other metric of interest). Thus, given an image $I$, the prediction $S_I$ and the reference database $\{J,S^{GT}_J\}_k$, we estimate the quality of $S_I$ as the average:

\begin{equation}
    DSC^{RCA}(I,S_i,\{J,S^{GT}_J\}_k) =  \frac{\sum_k DSC(\hat{S}_{J_k}, S^{GT}_{J_k})}{k}
    \label{eq:dsc-rca}
\end{equation} 

The underlying assumption is that if the segmentation quality of the new image is high, the RCA estimator will have a high performance on the reference images; conversely, poor segmentation quality will result in lower performance on the reference images. By aggregating the predictions of the RCA estimator on those images, one can assess the quality of the segmentation of a new image without requiring ground truth data. Note that even though RCA stands for reverse \textit{classification} accuracy, here there is no actual classifier, as the process is fully based on image registration. However, we keep the name following the original publication \cite{valindria2017reverse}.\\

\noindent \textbf{Deep registration networks for atlas-based RCA.} We propose a variant of the original atlas-based label propagation RCA framework which employs deep registration networks and top-$k$ image selection to reduce computational time. Our method has four main stages:
\begin{enumerate}
    \item First, the top-$k$ images, which are most similar to the atlas, are identified within our reference database, to avoid registering the atlas with all the references. 
    \item Subsequently, a deep CNN is employed to learn the affine registration parameters for every reference image. This network is implemented following a Siamese architecture, with two CNN encoders whose output are then processed by 2 fully connected layers to produce the affine parameters of a 2D transformation. 
    \item Then, a dense deformation field is estimated by an anatomically constrained deformable registration network (ACNN-RegNet) for every reference image which is finally used to warp the atlas mask (following Ref.~\cite{mansilla2020learning}).\footnote{Code for the full RCA pipeline based on deep registration networks is publicly available at \url{https://github.com/ngaggion/UBD_SourceCode}.}
    \item Finally, the estimated $DSC^{RCA}$ is computed following Eq. \ref{eq:dsc-rca}, i.e. by averaging over all estimates. Note that in the original work \cite{valindria2017reverse} the maximum is computed instead of the average of the DSC scores for the reference images. In our experiments, we found that taking the average was more robust than the maximum. We believe this difference may be due to the fact that we perform a top-$k$ selection where only the top-$k$ similar images are chosen, reducing the chance of having really different images in our reference set.
\end{enumerate}

\noindent \textbf{Unsupervised Bias Discovery.} Let us say we have a trained model \textbf{M} which, given an image $I$, produces a segmentation mask $S_I = \mathbf{M}(I)$. The idea is to audit \textbf{M} for fairness with respect to a demographic attribute \textbf{A}. Following previous work \cite{puyol2021fairness}, we will study fairness in terms of the DSC coefficient. In particular, we consider the difference in terms of DSC between demographic groups as a measure of bias (the closer to zero, the fairer the model). However, since we do not have access to segmentation masks for the target population, we will use the difference in terms of $DSC^{RCA}$ as a proxy for this performance gap. For example, let us assume $A$ refers to the biological sex. We then divide our target population according to whether they are male or female, and analyze the gap between the corresponding $DSC^{RCA}_{A=M}$ and $DSC^{RCA}_{A=F}$. Note that throughout the experimental section we will use the signed difference $\Delta DSC^{RCA} = DSC^{RCA}_{A=M} - DSC^{RCA}_{A=F}$ to be sure that our estimator is capturing the biases in the right direction. In what follows, we present synthetic and real experiments to validate the proposed UBD framework.

\section{Experiments and discussion}
We study the task of lung and heart segmentation in x-ray images, and measure fairness taking sex as the protected attribute. We employed four different x-ray datasets (comprising a total of 911 images) including JSRT \cite{jsrt_shiraishi2000development}, Montgomery \cite{montgomeryset}, Shenzhen \cite{shenzhen}, and a minor subset of the Padchest dataset \cite{bustos2020padchest} for which we had access to ground-truth annotations\footnote{For JSRT, Montgomery and Shenzhen we used the original annotations. For PadChest, we used the annotations released in the Chest X-ray Landmark Database \cite{Gaggion_2022} publicly available at \url{https://github.com/ngaggion/Chest-xray-landmark-dataset}}. 

All experiments were conducted using a UNet model \cite{ronneberger2015u} trained via a compound soft Dice and cross-entropy loss. In an effort to combine heterogeneous annotations --given that some images contain only lung annotations while others contain both lung and heart annotations, we adopted the approach outlined in \cite{gaggionISBI2023}. Here, each organ is treated as an independent output channel using binary versions of the loss functions, and gradients are back-propagated on a specific channel only when annotations for that channel are available.

\begin{figure}[t!]
    \centering
    \includegraphics[width=\linewidth]{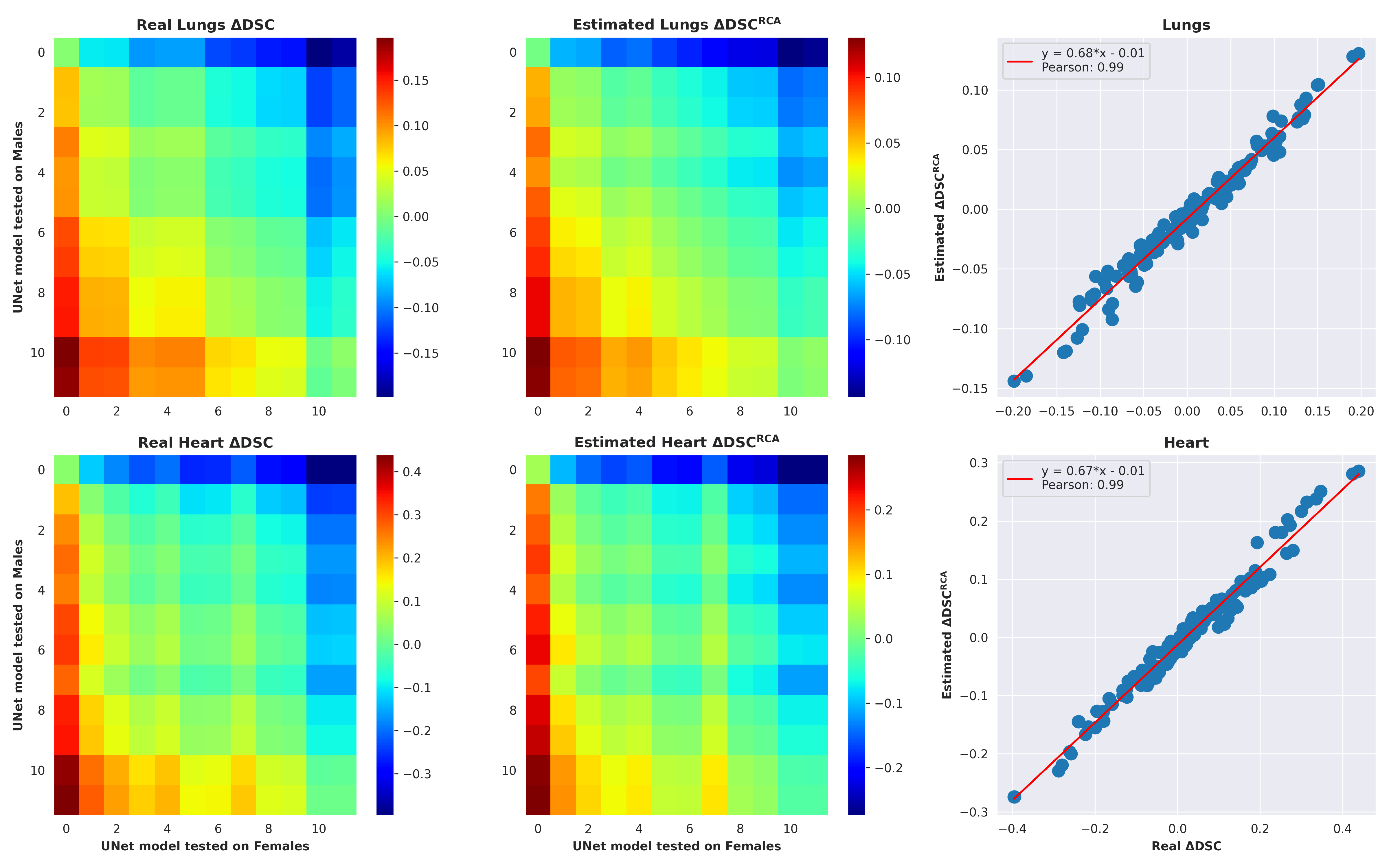}
    
    \caption{\textbf{Synthetic experiment.} Heatmaps illustrating the distribution of real bias as measured by $\Delta DSC$ (1st column) and the estimated $\Delta DSC^{RCA}$ (2nd column) across all combinations of models for male and female subjects. The third column shows scatter plots depicting the correlation between the real $\Delta DSC$ and the estimated $\Delta DSC^{RCA}$ bias. Each point represents a combination of models for male and female subjects. The positive correlation indicates that $\Delta DSC^{RCA}$ serves as a valid estimator of biases in the absence of ground-truth. Top/bottom row shows results for lung/heart segmentation.}
    \label{fig:bias_heatmaps}
\end{figure}

\subsection{Synthetic experiment: validating RCA for UBD}

We start with a synthetic experiment to validate the effectiveness of RCA as a bias detector. To obtain networks of varying performance levels, we trained UNet models until convergence, and used 12 different versions from intermediate saving checkpoints $M_{1\leq i \leq 12}$, one after each epoch from the 1st to the 11th and a final checkpoint at convergence. Note that since performance improves over training, in general the performance of model $M_i$ will tend to be higher than that of $M_j$ if $i>j$. We used balanced training and test sets in terms of sex (50\% male and 50\% female) (80-20\% split). To asses biases the test subset was analyzed in terms of the two sex-specific subgroups.

We simulated a scenario where the segmentation quality varies based on sex, with either male or female patients exhibiting superior performance. This was achieved by selecting pairs of UNet models $(M_i, M_j) \forall i,j : 1 \leq i,j \leq 12$, and segmenting the male test set with $M_i$ and the female set with $M_j$. Note that even though masks are coming from different models, in this synthetic experiment we consider them to be generated by a single fictitious model whose fairness we would aim to audit. This led to 144 possible combinations (fictitious models), where the segmentation quality in the male set was usually higher if $i>j$ and vice-versa. Figure \ref{fig:bias_heatmaps} (1st column) shows the real signed DSC gap for all possible UNet combinations tested on male ($M_i$) and female ($M_j$) patients, computed as $\Delta DSC = DSC_{A=M} - DSC_{A=F}$ (i.e. positive indicates higher segmentation quality for the male group, and negative for the female group). We then repeated the experiment but computing $\Delta DSC^{RCA}$, which does not require ground-truth annotations (see Figure \ref{fig:bias_heatmaps}, 2nd column). As it can be observed, $DSC^{RCA}$ is able to recover the same biases, up to a scaling factor. To better understand this relation, we also depict the real vs. estimated biases in a scatter plot shown in the 3rd column of Figure~\ref{fig:bias_heatmaps}.

Note that the direction of the bias, i.e., whether the performance is skewed towards males or females, can be inferred from the sign of the difference; positive values indicate a male bias and vice versa. We highlight that the signs of the $\Delta DSC$ and $\Delta DSC^{RCA}$  agree in 90\% and 92\% for the lungs and heart, respectively. When excluding cases where the bias was less than 0.01 (since small gaps may be due to statistical noise), the sign agreement improved to 96\% and 92\%. An even greater accuracy, 100\% and 96\%, was obtained when all pairs with bias less than 0.02 were excluded, thereby validating $\Delta DSC^{RCA}$ as a reliable estimator of bias in segmentation models.

\subsection{Real experiment: auditing chest x-ray segmentation models for sex bias}

\begin{figure}[t!]
    \centering
    \includegraphics[width=\linewidth]{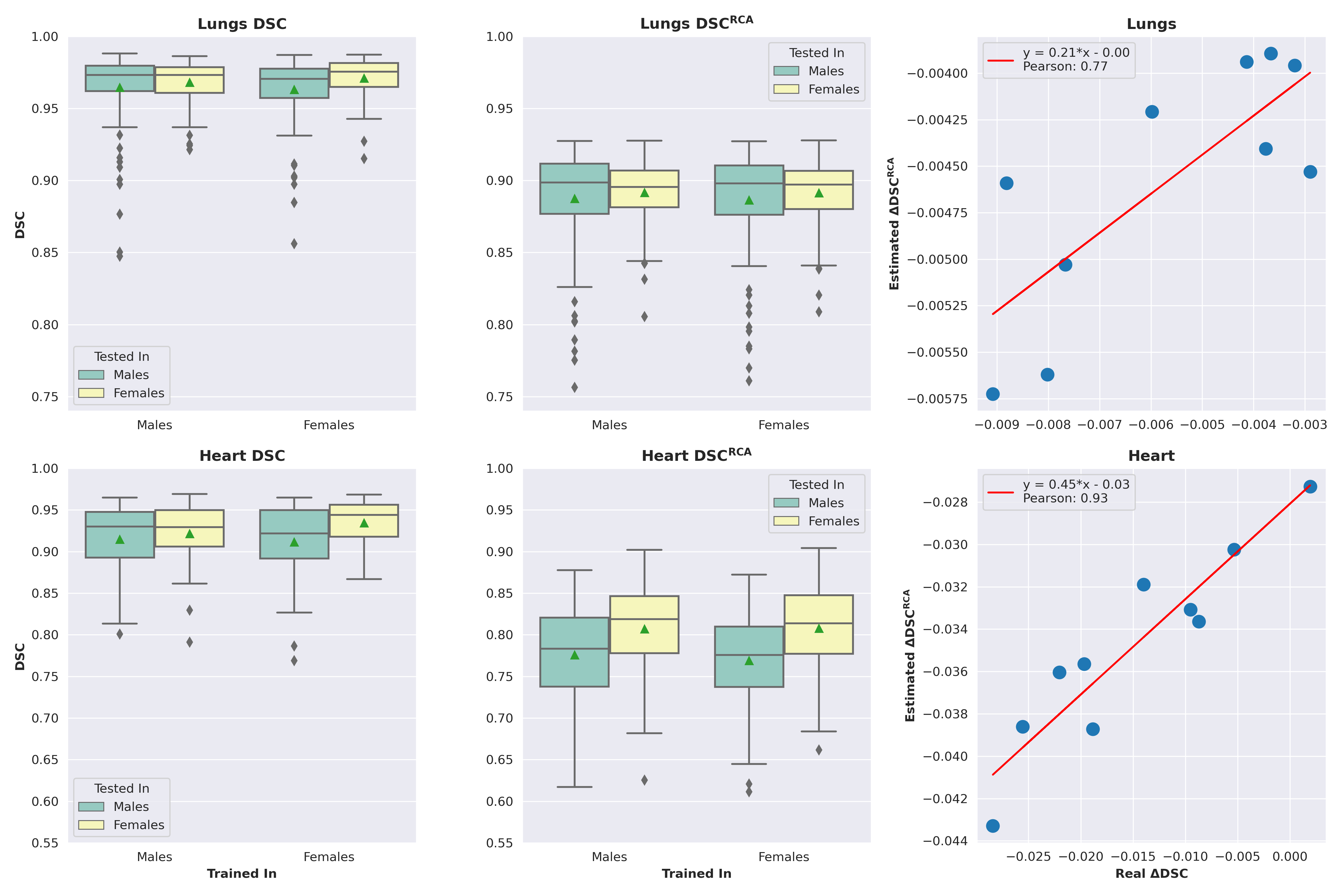}
    \caption{\textbf{Real experiment} Boxplots depict the distribution of real bias as measured by Real $DSC$ (1st column) and the estimated $DSC^{RCA}$ (2nd column) across all combinations of models trained and tested on male and female subjects separately. The third column shows scatter plots depicting the correlation between the real $\Delta DSC$ and the estimated $\Delta DSC^{RCA}$ bias. Each point represents a model. The positive correlation indicates that $\Delta DSC^{RCA}$ serves as a valid estimator of biases in the absence of ground-truth. Top/bottom row shows results for lungs/heart segmentation.}
    \label{fig:boxplots}
\end{figure}

We now proceed to validate $\Delta DSC^{RCA}$ in a more realistic scenario, auditing the bias of two concrete chest x-ray segmentation models. Data imbalance in terms of protected attributes in the training set is known to result in models which are biased in terms of those attributes \cite{larrazabal2020gender}. We exploit this fact to produce biased models by training two sets of 5-fold cross validation UNet models: one using only the male subset of the training set, and the other on the female subset. These models were then evaluated separately on the Male and Female partitions of the test set. Here again we employed an 80-20\% split for the complete train and test sets respectively. Intriguingly, when measuring biases using the available ground-truth annotations, we discovered that independently of the training set, models tend to perform better on female patients, with a slightly more pronounced gap when the model is trained on female patients (Figure \ref{fig:boxplots}, left and center column boxplots).

Crucially, estimated performance gaps via RCA strongly correlate with true gaps both for lung and heart segmentation also in this scenario (Figure~\ref{fig:boxplots}, right column top and bottom). However, it should be noted that the slope of the correlation is not exactly one and varies between tasks (cf. Figure~\ref{fig:boxplots} top vs bottom row). This suggests that while $\Delta DSC^{RCA}$ provides reliable estimates of bias up to a scaling factor, as in the synthetic experiment, calibration may be necessary to accurately estimate the magnitude of bias when transitioning from development to deployment scenarios.

\section{Conclusion}

The increasing adoption of automated methods in the context of MIC brings with it a responsibility to produce models which do not unfairly discriminate in terms of protected attributes. After development of a model, the reduced number of expert label annotations in the real-world dataset where the model is to be deployed may significantly limit the possibility to anticipate fairness issues. In order to estimate possible biases, here we propose a novel unsupervised bias discovery for biomedical image segmentation based on the RCA framework.

To evaluate our framework we employed x-ray datasets where ground truth annotations were actually available and true biases could hence be obtained. A first synthetic experiment, where the performance of a model could be easily manipulated, allowed us to validate our estimated bias score in a controlled scenario, showing the viability of our method. A second experiment performed in a more realistic scenario, where models were trained for high performance but in a data-imbalance setting resulting in biases, confirmed UBD was able to correctly capture these biases. Overall, these results show that UBD methods based on RCA could prove extremely helpful to anticipate biases at deployment in clinical settings where ground truth annotations are not yet available. 

\section{Acknowledgments}

This work was supported by Argentina’s National Scientific and Technical Research Council (CONICET), which covered the salaries of E.F., R.E. and D.M., and the fellowships of N.G. and L.M. The authors gratefully acknowledge NVIDIA Corporation with the donation of the GPUs used for this research, and the support of Universidad Nacional del Litoral (Grants CAID-PIC-50220140100084LI, 50620190100145LI), ANPCyT (PICT-PRH-2019-00009) and the Google Award for Inclusion Research (AIR) Program.

\bibliographystyle{splncs04}
\bibliography{refs}
\end{document}